\documentclass[10pt,twocolumn,letterpaper]{article}
\pdfoutput=1

\usepackage[pagenumbers]{cvpr} 

\usepackage[dvipsnames]{xcolor}

\usepackage{multirow}
\usepackage{booktabs}
\usepackage{ragged2e}
\usepackage{microtype}
\usepackage{color, colortbl}
\definecolor{Gray}{gray}{0.90}

\definecolor{cvprblue}{rgb}{0.21,0.49,0.74}
\usepackage[pagebackref,breaklinks,colorlinks,citecolor=cvprblue]{hyperref}

\def\paperID{5604} 
\def\confName{CVPR}
\def\confYear{2024}

\title{Video Anomaly Detection via Spatio-Temporal Pseudo-Anomaly Generation : \\ A Unified Approach}

\author{Ayush K. Rai$^1$, Tarun Krishna$^{*,1}$, Feiyan Hu$^{*,1}$, Alexandru Drimbarean$^2$ \\ Kevin McGuinness$^{\dagger,1}$, Alan F. Smeaton$^{\dagger,1}$, Noel E. O'Connor$^{\dagger,1}$\\
$^1$Insight SFI Centre for Data Analytics, Dublin City University $^2$Tobii Corporation, Galway \\
{\tt\small ayush.rai3@mail.dcu.ie}
}

\begin{document}
\maketitle

\def\thefootnote{*}\footnotetext{Equal contribution}\def\thefootnote{\arabic{footnote}}
\def\thefootnote{$\dagger$}\footnotetext{Equal supervision}\def\thefootnote{\arabic{footnote}}

\begin{abstract}

Video Anomaly Detection (VAD) is an open-set recognition task, which is usually formulated as a one-class classification (OCC) problem, where  training data is comprised of videos with  normal instances while test data contains both normal and anomalous instances. Recent works have investigated the creation of pseudo-anomalies (PAs) using only the normal data and making strong assumptions about real-world anomalies with regards to abnormality of objects and speed of motion to inject prior information about anomalies in an autoencoder (AE) based reconstruction model during training. This work proposes a novel method for generating generic spatio-temporal PAs by inpainting a masked out region of an image using a pre-trained Latent Diffusion Model and further perturbing the optical flow using mixup to emulate spatio-temporal distortions in the data. In addition, we present a simple unified framework to detect real-world anomalies under the OCC setting by learning three types of anomaly indicators, namely reconstruction quality, temporal irregularity and semantic inconsistency. Extensive experiments on four VAD benchmark datasets namely Ped2, Avenue, ShanghaiTech and UBnormal demonstrate that our method performs on par with other existing state-of-the-art PAs generation and reconstruction based methods under the OCC setting. Our analysis also examines the transferability and generalisation of PAs across these datasets, offering valuable insights by identifying real-world anomalies through PAs. Our results can be reproduced on   \href{https://github.com/rayush7/unified_PA}{\textcolor{red}{github}}.


\end{abstract}    
\section{Introduction}
\label{sec:intro}

\begin{figure*}[t]
    \centering
    \includegraphics[width=0.9\textwidth]{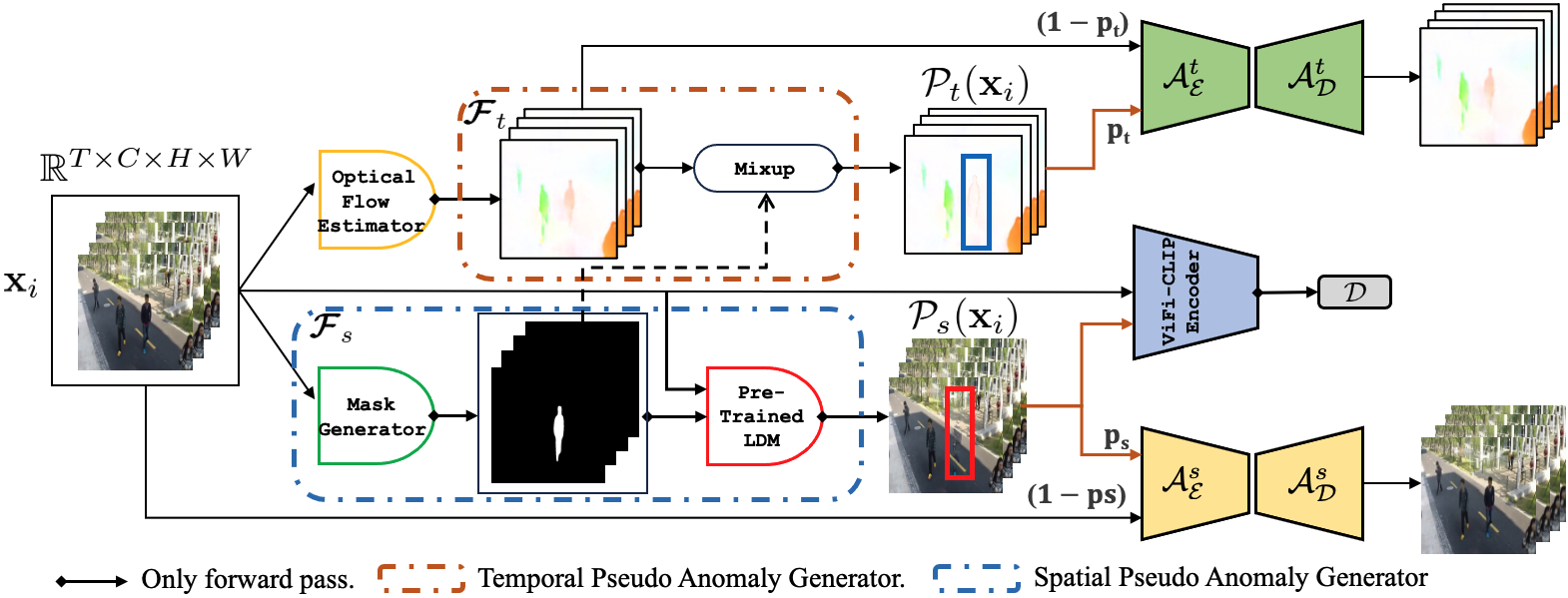}
    \caption{The overall architecture of our approach consists of spatio-temporal PAs generators. Spatial PAs generator (eq. \ref{spatial_pseudo_anomaly_eq})~:~$\mathcal{F}_{s}( \text{stack}(\mathbf{x}, \mathbf{x} \odot \mathbf{m}, \mathbf{m}); \theta)$and temporal PAs (eq. \ref{temporal_pseudo_anomaly_eq})~:~$\mathcal{F}_{t} (\phi (\mathbf{x_{t}}, \mathbf{x_{(t+1)}}))$. The spatial and temporal PAs are sampled with probability $p_s$ and $p_t$ respectively. Our VAD framework unifies estimation of reconstruction quality (eq. \ref{spatial_loss_objective}), temporal irregularity (eq. \ref{temporal_loss_objective}) and semantic inconsistency.}
    \label{fig:arch}
\end{figure*}

Video Anomaly Detection \cite{liu2018future,liu2021hybrid,ionescu2019object,zaheer2020old,Gong_2019_ICCV,park2020learning,astrid2021learning,astrid2021synthetic,sultani2018real,Pourreza_2021_WACV,georgescu2021background,georgescu2021anomaly,ji2020tam,wang2022video,zaheer2022generative} refers to the task of discovering the unexpected occurrence of events that are distinct and follow a deviation from known normal patterns. 
The rarity of anomalies in the real-world and the unbounded nature (open-set recognition \cite{geng2020recent}) of their diversities and complexities have led to unbalanced training datasets for VAD making it an extremely challenging task. Therefore VAD is commonly addressed as a OCC problem where only normal data is available to train a model \cite{hasan2016learning,zhao2017_acmmm,Luo_2017_ICCV,8019325, Gong_2019_ICCV,park2020learning,astrid2021learning,astrid2021synthetic,georgescu2021anomaly,liu2021hybrid}.

Reconstruction based approaches exploiting an AE are usually adopted to tackle the OCC task \cite{astrid2021learning,astrid2021synthetic,park2020learning,Gong_2019_ICCV}. The intuition behind this is that during training, the AE would learn to encode normal instances in its feature space with the assumption that during the test phase a high reconstruction error would correspond to an anomaly and a low reconstruction error would indicate normal behaviour. Contrary to this, \cite{Gong_2019_ICCV,astrid2021learning,zaheer2020old} observed that when trained in this setting,  the AE learns to reconstruct anomalies with high accuracy resulting in a low reconstruction error in the testing phase. Hence, the capability of the AE to distinguish normal and anomalous instances is greatly diminished (Figure~1a in \cite{astrid2021learning}).

\cite{park2020learning,Gong_2019_ICCV} introduced a memory-based AE to restrict the reconstruction capability of the AE by recording  prototypical normal patterns during training in the latent space therefore shrinking the capability of the AE to reconstruct anomalous data. However, such methods are highly sensitive to  memory size. A small-sized memory may hinder reconstruction of  normal data as  memorising normal patterns can be interpreted as severely limiting the reconstruction boundary of the AE, resulting in failure to reconstruct even the normal events during the testing phase (Figure~1b in \cite{astrid2021learning}). 

Astrid \etal \cite{astrid2021learning} proposed the generation of two types of PAs (patch based and skip-frame based) to synthetically simulate pseudo-anomalous data  from normal data and further introduced a novel training objective for the AE to force the reconstruction of only normal data even if the input samples are anomalous. Patch based PAs are generated by inserting a patch of a specific size and orientation from an intruder dataset (e.g. CIFAR-100) using the  SmoothMixS \cite{Lee_2020_CVPR_Workshops} data augmentation method  while in order to create skip-frame based PAs, a sequence of frames is sampled with irregular strides to create anomalous movements in the sequence. The intuition behind this training procedure is based on limiting the reconstruction boundary of the AE near the boundaries of the normal data resulting in more distinctive features between normal and anomalous data (Figure~1c in \cite{astrid2021learning}). A notable limitation of the approach proposed in Astrid \etal \cite{astrid2021learning} is its heavy reliance on a predefined set of assumptions and inductive biases. These assumptions encompass various aspects, including the specific intruding dataset selected for patch insertion, the patch's size and orientation, and the idea that altering the movement speed by skipping frames could introduce temporal irregularities into the normal data. 


With such assumptions, there is no guarantee that the test anomalies which comprise of an unbounded set of possible anomalous scenarios would comply with pseudo-anomalous samples. This creates a need for more generic solutions for creating PAs from the normal data. Since VAD is an open-set recognition problem and anomalies present an inexhaustible set of possibilities, every pseudo-anomaly synthesiser carries strong or weak inductive biases and thus it is inherently challenging  to emulate real anomalies through PAs. Furthermore, there are other challenges, such as the fact that certain normal behaviours are rare but possible and therefore not well represented in the normal data. This presents an interesting research question: \textit{``Is it possible to synthetically generate generic PAs by introducing spatio-temporal distortions into normal data in order to detect real-world anomalies effectively?, and importantly, can such PAs transfer across multiple VAD datasets?''} 

Our work is motivated by \cite{astrid2021learning} and extends it by addressing its drawbacks  and proposing a more generic pseudo-anomaly generator. We focus on generating PAs by injecting two different types of anomaly indicators, the first being distortion added through image inpainting  performed by a pre-trained latent diffusion model (LDM) \cite{rombach2022high}, the second being the addition of temporal irregularity through perturbation of the optical flow \cite{zach2007duality} using mixup \cite{zhang2017mixup}. In addition, our method also measures the semantic inconsistency between normal samples and PAs using semantically rich ViFi-CLIP \cite{rasheed2023fine} features. This \textit{unifies estimation of reconstruction quality, temporal irregularity and semantic inconsistency} under one framework. We conduct an extensive study on understanding the generalisation and transferability of such PAs over real-world anomalies. Overall, our main contributions are:
\begin{itemize}
    \item We propose a novel and generic spatio-temporal pseudo-anomaly generator for VAD encompassing inpainting of a masked out region in  frames using an LDM and applying mixup augmentation to distort the optical flow. 

    
    \item We introduce a unified VAD framework that measures and aggregates three different indicators of anomalous behaviour namely reconstruction quality, temporal irregularity and semantic inconsistency in an OCC setting.
    
    \item Extensive experiments on 
    \textit{Ped2, Avenue, ShanghaiTech} and \textit{UBnormal} show that our method achieves on par performance with other existing SOTA methods (Table~\ref{results_sota}, \ref{results_ubnormal}) indicating that our method is a generic video anomaly detector and our spatio-temporal pseudo anomaly generation process is transferable across multiple datasets.
\end{itemize}
\section{Related Work}
\label{sec:related_work}

\subsection{Restricting Reconstruction Capacity of an AE}
A standard approach to address VAD is to adopt an OCC strategy by training an AE model to reconstruct the input data \cite{hasan2016learning,zhao2017_acmmm,Luo_2017_ICCV,8019325,Gong_2019_ICCV,park2020learning,astrid2021learning}. During training, only normal inputs are used for learning the AE with the assumption that reconstruction of anomalies during testing would yield a higher reconstruction error. However,  in practice it has been shown that the AE can also reconstruct  anomalous data  \cite{Gong_2019_ICCV,astrid2021learning,zaheer2020old}. \cite{Gong_2019_ICCV,park2020learning} mitigated this issue by augmenting the AE with memory-based techniques in the latent space to restrict the reconstruction capability of an AE. However the performance of such methods are directly impacted by the choice of the memory size, which may over-constrain the reconstruction power of the AE resulting in poor reconstruction of even the normal events during testing.

To alleviate this issue, \cite{astrid2021learning,astrid2021synthetic} utilised data-heuristic based PAs built on strong assumptions to limit the reconstruction capacity of the AE. Patch-based PAs were generated by inserting a patch from an intruding dataset (CIFAR-100) into the normal data by using techniques such as SmoothMixS \cite{Lee_2020_CVPR_Workshops}. For modeling motion-specific anomalous events, PAs were generated by skipping frames with different strides to induce temporal irregularity. The training configuration was set up to minimise the reconstruction loss of the AE with respect to the normal data only. PAs can be interpreted as a type of data-augmentation  \cite{pmlr-v15-bengio11b,krizhevsky2012imagenet}, where instead of creating more data of the same distribution, pseudo-anomalous data is created that belongs to a near-distribution i.e. between the normal and anomaly distributions. \cite{tang2020onlineaugment,zhang2019adversarial} adopted adversarial training to generate augmented inputs, which were also effective as an adversarial example for the model. 

Our method  falls into the category of restricting the reconstruction capability of an AE, where  we follow the training setup introduced in \cite{astrid2021learning}, however we propose simulation of generic spatio-temporal PAs without making bold assumptions about dataset specific anomalies.


\subsection{Generative Modeling} 

Generative models have been used to generate out of distribution (OOD) data for various applications in semi-supervised learning (Bad GAN \cite{dai2017good}, Margin GAN \cite{dong2019margingan}), anomaly detection (Fence GAN \cite{ngo2019fence}), OOD detection (BDSG \cite{dionelis2020boundary,du2022vos}), medical anomaly detection \cite{wolleb2022diffusion} and novelty detection \cite{mirzaeifake}. However,  such methods mostly work with low dimensional data and are not suitable  for generating OOD data for VAD. OGNet \cite{zaheer2020old,zaheer2022stabilizing} and G2D \cite{Pourreza_2021_WACV} exploit a GAN-based generator and discriminator for VAD. During the first phase of training, a pre-trained state of the generator is used to create  PAs or irregular samples while in the second phase, binary classification is performed using a discriminator to distinguish between normal and PAs samples.

We design our model from the perspective of generating generic spatio-temporal PAs where a generative model (pre-trained LDM) is availed to generate spatial PAs while the mixup method is exploited to create temporal PAs from optical flow.


\subsection{Other VAD Methods}
\noindent \textbf{Non-Reconstruction Based Methods:} Several non-reconstruction based methods have also been proposed which derive their anomaly scores from various different indicators of anomaly in addition to reconstruction loss. The work presented in \cite{liu2018future} utilised a future frame prediction task for VAD and estimated optical flow and gradient loss as supplementary cues for anomalous behaviour. \cite{georgescu2021anomaly,ionescu2019object} performed object detection as a pre-processing step under the assumption that anomalous events are always object-centric. Several other works added optical flow components \cite{ji2020tam,bman} to detect anomalous motion patterns and a binary classifier \cite{zaheer2020old,Pourreza_2021_WACV} to estimate anomaly scores. 

\noindent

In our work, we also use a segmentation mask (object detection) and optical flow to generate corresponding spatial and temporal PAs during the training phase. However during inference we do not carry out any object detection and perform anomaly detection solely based on reconstruction of images and on optical flow. 

\noindent \textbf{Non-OCC methods:}
There are many other formulations of VAD. \cite{georgescu2021anomaly} introduced a self-supervised method where different pretext tasks such as arrow of time, middle-box prediction, irregular motion discrimination and knowledge distillation were jointly optimised for VAD. \cite{wang2022video} adopted a self-supervised single pre-text task of solving decoupled temporal and spatial jigsaw puzzles corresponding to modeling normal appearance and motion patterns. Several works have also addressed the VAD problem as a weakly supervised problem through multiple instance learning
\cite{sultani2018real,wu2020not, zhu2022towards,zhang2023exploiting}. Unsupervised VAD methods involve the cooperation of two networks through an iterative process for pseudo-label generation \cite{zaheer2022generative,pang2020self,zaheer2020claws,cleaning2020zaheer,zaheer2020self,lin2022causal}. Zero-shot VAD was introduced in \cite{aich2023cross} where a model was trained on the source domain to detect anomalies in a target domain without any domain adaptation. USTN-DSC \cite{yang2023video} a proposed video event restoration framework based on keyframes for VAD while EVAL \cite{singh2023eval} presented a technique for video anomaly localisation allowing for human interpretable explanations. 

\begin{figure}[h!]
    \centering
    \includegraphics[width=0.49\textwidth]{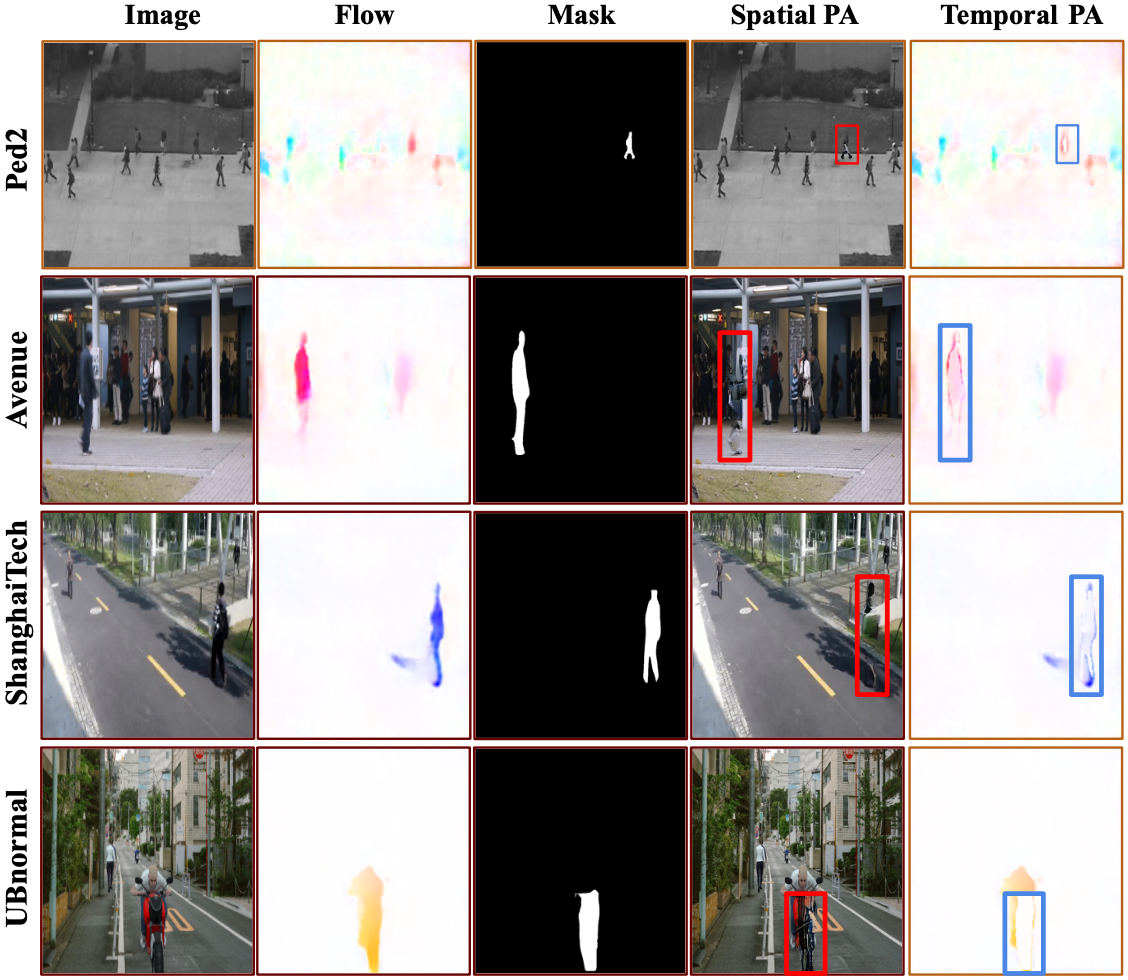}
    \caption{
    Visualisation of spatial and temporal PAs, using segmentation masks. This approach also works with random masks.}
    \label{fig:pseudo_ano}
\end{figure}
\section{Method}
\label{sec:method}


\subsection{Preliminaries}
\noindent \textbf{Latent Diffusion Models (LDMs):} Diffusion Probabilistic Models (DMs) \cite{sohl2015deep,ho2020denoising,song2020score} are a class of probabilistic generative models that are designed for learning a data distribution $p_{\text{data}}(\mathbf{x})$. DMs iteratively denoise a normally distributed variable by learning the reverse process of a fixed Markov Chain of length T through a denoising score matching objective~\cite{song2020score} given by:
\begin{equation}
    \label{diff_model_eq}
    \mathbb{E}_{\mathbf{x} \sim p_{\text{data}}, \tau \sim p_{\tau}, \epsilon \sim \mathcal{N}(\mathbf{0},\mathbf{I})} [|| \mathbf{y} - \mathbf{f}_{\theta} (\mathbf{x}_{\tau};\mathbf{c},\tau) ||^2_{2}],
\end{equation}
 where $\mathbf{x} \sim p_{data}$, the diffused input can be constructed by $\mathbf{x_{\tau}} = \alpha_{\tau} \mathbf{x} + \sigma_{\tau} \mathbf{\epsilon}$, $\epsilon \sim \mathcal{N}(\mathbf{0},\mathbf{I})$ and is fed into a denoiser model $\mathbf{f_{\theta}}$, 
 $(\sigma_{\tau}, \alpha_{\tau})$ denotes the noise schedule parameterised by diffusion-time $\tau$, $p_{\tau}$ is a uniform distribution over $\tau$, $\textbf{c}$ denotes conditioning information and the target vector $\mathbf{y}$ is either the random noise $\epsilon$ or $ \mathbf{v} = \alpha_{\tau}\epsilon - \sigma_{\tau} \mathbf{x}$. The forward diffusion process corresponds to gradual addition of the gaussian noise to $\mathbf{x}$ such that the logarithmic signal-to-noise ratio $\lambda_{\tau} = \log(\alpha_{\tau}^2/ \sigma_{\tau}^2)$ monotonically decreases.

LDMs  \cite{rombach2022high} were proposed to make standard DMs efficient by training a VQGAN \cite{esser2021taming} based model to project input images i.e. $\mathbf{x} \sim p_{data}$ into a spatially lower dimensional latent space of reduced complexity and then reconstructing the actual input with high accuracy. In particular, a regularised AE \cite{rombach2022high} is used to reconstruct the input $\mathbf{x}$ such that the reconstruction is given by : $\mathbf{\hat{x}} = \mathbf{f_{de}} \circ \mathbf{f_{en}}(\mathbf{x}) \footnote{ $\circ:$ denotes function composition} \thickapprox \mathbf{x}$, where $\mathbf{f_{en}}$ and $\mathbf{f_{de}}$ denotes encoder and decoder respectively. Furthermore an adversarial objective is added using a patch-based discriminator \cite{isola2017image} to ensure photorealistic reconstruction. DM is then trained in the latent space by replacing $\mathbf{x}$ with its latent representation $\mathbf{z} = \mathbf{f_{en}}(\mathbf{x})$ in eq.~(\ref{diff_model_eq}).
This leads to reduction in \# of learnable parameters and memory. 

\subsection{Generating Spatial-PAs}
\noindent Real world anomalies are highly context specific without having a ubiquitous definition. Ramachandra~\textit{et al.}~\cite{ramachandra2020survey} loosely define them as, the ``occurrence of unusual appearance and motion attributes or the occurrence of usual appearance and motion attributes at an unusual locations or times''. Examples of such cases include: an abandoned object in a crowded area or suspicious behaviour of an individual. We address this notion of occurrence of unusual appearance attributes through generation of spatial PAs. 

Since LDMs  achieve state-of-the-art performance for the image inpainting task,  this can be exploited as a spatial PAs generator. In particular, we hypothesise that an off-the-shelf pre-trained LDM model \cite{rombach2022high} without any finetuning on VAD datasets can inpaint the image with enough spatial distortion that can serve as spatially pseudo-anomalous samples for training a VAD model. We follow the mask generation strategy proposed in LAMA \cite{suvorov2022resolution} to generate both randomly shaped and object segmentation masks $\mathbf{m}$. We concatenate image $\mathbf{x}$, masked image $\mathbf{x} \odot \mathbf{m}$ \footnote{ $\odot:$ denotes point-wise multiplication} and mask $\mathbf{m}$ over the channel dimension and give this 7 channel input to UNet \cite{ronneberger2015u}.
We denote the normal data samples as $\mathbf{x}$ unless otherwise explicitly stated. The spatial PAs $\mathcal{P}_s(\mathbf{x})$ is given by:


\begin{equation}
   \label{spatial_pseudo_anomaly_eq}
   \mathcal{P}_s(\mathbf{x})= \mathcal{F}_{s}( \text{stack}(\mathbf{x} , \mathbf{x} \odot \mathbf{m}, \mathbf{m}); \theta),
\end{equation}


\noindent 
where $\mathcal{F}_{s}$ is the inpainting model that uses latent diffusion with pre-trained model parameters $\theta$. Some examples of the spatial PAs are shown in Figure~\ref{fig:pseudo_ano}.




\subsection{Generating Temporal-PAs}
We address the notion of unusual motion occurrences (such as person falling to ground) through the generation of temporal PAs. 
Various video diffusion models \cite{he2022latent,ho2022video,voleti2022mcvd} have been proposed, which can be exploited to induce temporal irregularity in the video.  
However due to limited computational resources, we introduce a simple but effective strategy for the generation of temporal PAs by applying  a vicinal risk minimisation technique mixup \cite{zhang2017mixup} to the optical flow of the normal videos.
More specifically, given a \textit{normal} video $\mathbf{v}$, its frame $\mathbf{x_{t}}$, and  its corresponding segmentation mask $\mathbf{m_{t}}$ and another consecutive frame $\mathbf{x_{(t+1)}}$, we compute the optical flow $\mathbf{\phi}(\mathbf{x_{t}},\mathbf{x_{(t+1)}})$ using the TVL1 alogrithm \cite{zach2007duality}. For simplification, we use $\mathbf{\phi}$ as an alias to represent $\mathbf{\phi}(\mathbf{x_{t}},\mathbf{x_{(t+1)}})$. Let us consider a rectangular patch  $\mathbf{p}^\prime$ in $\mathbf{\phi}$ corresponding to the mask $\mathbf{m_{t}}$ in the frame $\mathbf{x_{t}}$ with dimensions $\mu_{h}$ and $\mu_{w}$. In order to perturb the optical flow $\mathbf{\phi}$, we take another rectangular patch $\mathbf{p_r}^\prime$ at a random location in $\mathbf{\phi}$ with the same dimensions as $\mathbf{p}^\prime$ and apply mixup to yield $\mathbf{\hat{p}}$, which is a convex combination of $\mathbf{p}^\prime$ and $\mathbf{p_r}^\prime$  given by : $\mathbf{\hat{p}} = \lambda \mathbf{p}^\prime + (1 - \lambda) \mathbf{p_r}^\prime$, where $\lambda$ is sampled from a beta distribution with $\alpha=0.4$ as in \cite{zhang2017mixup}.
We denote the temporal PAs as $\mathcal{P}_t(\mathbf{x})$ given by: 




\begin{equation}
   \label{temporal_pseudo_anomaly_eq}
   \mathcal{P}_t(\mathbf{x}) = \mathcal{F}_{t} (\phi (\mathbf{x_{t}}, \mathbf{x_{(t+1)}})),
\end{equation}


\noindent 
where $\mathcal{F}_{t}$ is the temporal PAs generator. Some examples of  temporal PAs are depicted in Figure~\ref{fig:pseudo_ano}. It is important to note that our PAs generation method does not explicitly require segmentation masks, it can also generate PAs using random masks. Since segmentation masks carry semantic meaning, using them enables generation of more semantically informative PAs as further validated by our experiments. 

\subsection{Reconstruction Model}
The training mechanism follows a similar strategy as in \cite{astrid2021learning}, where regardless of the input ($\mathcal{I}$) i.e normal ($\mathbf{x}$/$\mathbf{\phi}$) or PAs ($\mathcal{P}_s(\mathbf{x})$/$\mathcal{P}_t(\mathbf{x})$) the network is forced to reconstruct only the normal input using a 3D-CNN (Convolutional Neural Network) based AE model adapted from the convolution-deconvolution network proposed by \cite{Gong_2019_ICCV} (Table~\ref{model_arch} in supplementary material (supp.)).

We train two different AEs with the aim of limiting their reconstruction capacity by exposing them to spatial and temporal PAs. We represent the spatial (temporal) AE by $\mathcal{A}^{s} (\mathcal{A}^{t})$ with $\mathcal{A}_\text{e}^{s} (\mathcal{A}_\text{e}^{t})$  and $\mathcal{A}_\text{de}^{s} (\mathcal{A}_\text{de}^{t})$ denoting its encoder and decoder respectively. The reconstruction output of $\mathcal{A}^{s}$ is given by : $\mathbf{\hat{x}} = \mathcal{A}_\text{de}^{s} \circ \mathcal{A}_\text{e}^{s} (\mathbf{x})$ while the reconstruction output of $\mathcal{A}^{t}$ is computed by : $\mathbf{\hat{\phi}} = \mathcal{A}_\text{de}^{t} \circ \mathcal{A}_\text{e}^{t} (\mathbf{\phi})$. In order to train $\mathcal{A}^{s}$ and $\mathcal{A}^{t}$, PAs ($\mathcal{P}_s(\mathbf{x})$ or $\mathcal{P}_t(\mathbf{x})$) are given as respective inputs with a probability $p_s$ (or $p_t$) while the normal data is provided as input with probability of $(1-p_s)$ (or $(1-p_t)$). $p_s$ (or $p_t$) is a hyperparameter to control the ratio of PAs to normal samples. Overall, the loss for $\mathcal{A}^{s}$ and $\mathcal{A}^{t}$ is calculated as:


\begin{equation}
\label{spatial_loss_objective}
\mathcal{L}_{\mathcal{A}^{(s)}} =\frac{1}{\Pi}
\begin{cases}
    || \mathbf{\hat{x}} - \mathbf{x} ||^{2}_{2} & \text{if } \mathcal{I} = \mathbf{x}  \\
    || \mathcal{\hat{P}}_s(\mathbf{x}) - \mathbf{x} ||^{2}_{2}  & \text{if } \mathcal{I} = \mathcal{P}_s(\mathbf{x}), \\
\end{cases}
\end{equation}

\begin{equation}
\label{temporal_loss_objective}
\mathcal{L}_{\mathcal{A}^{(t)}} = \frac{1}{\Pi}
\begin{cases}
    || \mathbf{\hat{\phi}} - \mathbf{\phi} ||^{2}_{2} & \text{if } \mathcal{I} = \mathbf{\phi}  \\
    || \mathcal{\hat{P}}_t(\mathbf{x}) - \mathbf{\phi} ||^{2}_{2}  & \text{if } \mathcal{I} = \mathcal{P}_t(\mathbf{x}), \\
\end{cases}
\end{equation}

\noindent 
where $1/\Pi$ is normalisation factor,  $\Pi = \mathcal{T} \times C \times H \times W$ and $||.||_{2}$ denotes the $\mathcal{L}_2$ norm. where $\mathcal{T}, C, H \text{ and } W$ are  the number of  frames, number of channels, height, and width of the frames in the input sequence ($\mathcal{I}$), respectively.

\subsection{Estimating Semantic Inconsistency}
While measuring the spatial reconstruction quality and temporal irregularity between normal and anomalous data is essential for real-world VAD, it is also  crucial to learn and estimate the semantic inconsistency (degree of misalignment of semantic visual patterns and cues) between normal and anomalous samples (e.g. abnormal object in the crowded scene). In practice, to emulate this idea in our approach, we extract frame-level semantically rich features from the ViFi-CLIP \cite{rasheed2023fine} model (pre-trained on Kinetics-400 \cite{kay2017kinetics}) and perform binary classification between normal data samples $\mathbf{x}$ and spatial pseudo-anomalies $\mathcal{P}_s(\mathbf{x})$ using a discriminator $\mathcal{D}$, 
(Table~\ref{disc_arch} in supp.),
which can be viewed as an auxiliary component to AEs. Intuitively, it is highly likely that latent space representation of PAs will be semantically inconsistent to the normal scenarios. 

\begin{table*}[h]
\caption{Micro AUC score comparison between our approach and state-of-the-art methods on test split of Ped2 \cite{ped2}, Avenue (Ave) \cite{avenue_6751449} and  ShanghaiTech (Sh) \cite{luo2017revisit}. Best and second best performances are
highlighted as \textbf{bold} and \underline{underlined}, in each category and dataset.}
\label{results_sota}
\resizebox{\linewidth}{!}{
\centering
\begin{tabular}[t]{|c|l|ccc|}
\hline
\multicolumn{2}{|c|}{Methods}                  & Ped2 \cite{ped2}  & Ave \cite{avenue_6751449} & Sh \cite{luo2017revisit}     \\ \hline \hline
\parbox[t]{2mm}{\multirow{18}{*}{\rotatebox[origin=c]{90}{Miscellaneous}}}           
  & OLED \cite{Jewell_2022_WACV} & \underline{99.02\%} & - & - \\
  & AbnormalGAN \cite{ravanbakhsh2017abnormal}     & 93.50\%  & -       & -       \\
  & Smeureanu \etal \cite{smeureanu2017deep}       & -       & 84.60\%  & -       \\
  & AMDN \cite{xu2015learning,xu2017detecting}                    & 90.80\%  & -       & -       \\
  & STAN \cite{lee2018stan}                        & 96.50\%  & 87.20\%  & -       \\
  & MC2ST \cite{liu2018classifier}                 & 87.50\%  & 84.40\%  & -       \\
  & Ionescu \etal   \cite{ionescu2019detecting}    & -       & 88.90\%  & -       \\
  & BMAN \cite{lee2019bman}                        & 96.60\%  & \textbf{90.00\%}  & 76.20\%  \\
  & AMC  \cite{nguyen2019anomaly}                  & 96.20\%  & 86.90\%  & -       \\
  & Vu \etal \cite{vu2019robust}                   & \textbf{99.21\%}  & 71.54\%  & -       \\
  & DeepOC   \cite{wu2019deep}                     & -     & 86.60\%  & -       \\
  & TAM-Net  \cite{ji2020tam}                      & 98.10\%  & 78.30\%  & -       \\
  & LSA \cite{abati2019latent}                     & 95.40\%  & -       & 72.50\%  \\
  & Ramachandra \etal \cite{ramachandra2020learning} & 94.00\%  & 87.20\%  & -   \\
  & Tang \etal \cite{tang2020integrating}          & 96.30\%  & 85.10\%  & 73.00\%  \\
  & Wang \etal \cite{wang2020cluster}              & -       & 87.00\%  & \textbf{79.30\%}  \\
  & OGNet \cite{zaheer2020old}                     & 98.10\%  & -       & -       \\
  & Conv-VRNN \cite{lu2019future}         & 96.06\%  & 85.78\%  & -  \\ 
  & Chang \etal \cite{chang2020clustering}         & 96.50\%  & 86.00\%  & 73.30\%  \\ 
  & USTN-DSC \cite{yang2023video} & 98.10\% & \underline{89.90\%} & 73.8\%  \\ 
  & EVAL \cite{singh2023eval} & - & 86.06\% & \underline{76.63\%} \\ \hline
\parbox[t]{2mm}{\multirow{6}{*}{\rotatebox[origin=c]{90}{Object-centric}}}   
  & MT-FRCN \cite{hinami2017joint}                     & 92.20\%  & -       & -       \\
  & Ionescu \etal \cite{ionescu2019object} \footnotemark             & 94.30\%  & 87.40\%  & 78.70\%  \\
  & Doshi and Yilmaz \cite{doshi2020any,doshi2020continual} & \underline{97.80\%}  & 86.40\%  & 71.62\%  \\
  & Sun \etal \cite{sun2020scene}                      & -       & 89.60\% & 74.70\%  \\
  & VEC \cite{yu2020cloze}                             & 97.30\%  & \underline{90.20\%}  & \underline{74.80\%}  \\ 
  & Georgescu \etal \cite{georgescu2021background}             & \textbf{98.70\%}  & \textbf{92.30\%}  & \textbf{82.70\%} \\ \hline

\end{tabular}
\begin{tabular}[t]{|c|l|ccc|}
\hline
\multicolumn{2}{|c|}{Methods}                  & Ped2 \cite{ped2}  & Ave \cite{avenue_6751449} & Sh \cite{luo2017revisit}     \\ \hline \hline
\parbox[t]{2mm}{\multirow{7}{*}{\rotatebox[origin=c]{90}{Non deep learning}}}
  & MDT   \cite{mahadevan2010anomaly}                  & 82.90\%  & -       & -     \\
  & Lu \etal \cite{avenue_6751449}                     & -       & \textbf{80.90\%}  & -     \\
  & AMDN   \cite{xu2017detecting}                      & \underline{90.80\%}  & -       & -     \\
  & Del Giorno \etal \cite{del2016discriminative}      & -       & \underline{78.30\%}  & -     \\
  & LSHF    \cite{zhang2016video}                      & \textbf{91.00\%}  & -       & -     \\
  & Xu \etal \cite{xu2014video} & 88.20\%  & -  & -     \\
  & Ramachandra and Jones \cite{ramachandra2020street} & 88.30\%  & 72.00\%  & -     \\ \hline
  \parbox[t]{2mm}{\multirow{7}{*}{\rotatebox[origin=c]{90}{Prediction}}}        
  & Frame-Pred  \cite{liu2018future}          & 95.40\%  & 85.10\%  & 72.80\%  \\
  & Dong \etal \cite{dong2020dual}            & 95.60\%  & 84.90\%  & 73.70\%  \\
  & Lu \etal \cite{lu2020few}                 & 96.20\%  & 85.80\%  & \textbf{77.90\%}  \\
  & MNAD-Pred \cite{park2020learning}         & \underline{97.00\%}  & \underline{88.50\%}  & 70.50\%  \\ 
  & AnoPCN \cite{ye2019anopcn}                & 96.80\% & 86.20\% & 73.60\% \\
  & AMMC-Net \cite{ammc-net}                  & 96.90\% & 86.60\% & 73.70\%\\
  & DLAN-AC \cite{yang2022dynamic} & \textbf{97.60\%} & \textbf{89.90\%} & \underline{74.70\%} \\
  
  \hline
\parbox[t]{2mm}{\multirow{12}{*}{\rotatebox[origin=c]{90}{Reconstruction}}}   
  & AE-Conv2D  \cite{hasan2016learning}          & 90.00\%  & 70.20\%  & 60.85\% \\
  & AE-Conv3D  \cite{zhao2017spatio}             & 91.20\%  & 71.10\%  & -       \\
  & AE-ConvLSTM  \cite{luo2017remembering}       & 88.10\% & 77.00\% & -       \\
  & TSC \cite{luo2017revisit}                    & 91.03\% & 80.56\% & 67.94\% \\
  & StackRNN \cite{luo2017revisit}               & 92.21\% & 81.71\% & 68.00\% \\
  & MemAE \cite{gong2019memorizing}              & 94.10\%  & 83.30\%  & 71.20\%  \\
  & MNAD-Recon \cite{park2020learning}           & 90.20\%  & 82.80\%  & 69.80\%  \\ 
  & \textit{Baseline} (without PAs)             & 92.49\%  & 81.47\% & 71.28\% \\
  & STEAL Net \cite{astrid2021synthetic} & \textbf{98.40\%} & \textbf{87.10\%} & \underline{73.70\%} \\
  & LNTRA Astrid \etal \cite{astrid2021learning}  - Patch based      &  94.77\% & 84.91\% & 72.46\% \\
  & LNTRA Astrid \etal \cite{astrid2021learning} - Skip-frame based      & \underline{96.50\%} & 84.67\% & \textbf{75.97\%} \\ 
  \rowcolor{Gray}
  & Ours w/o $\mathcal{D}$ & 93.52\% & 86.51\% & 71.76\% \\
  \rowcolor{Gray}
  & Ours w/ $\mathcal{D}$ & 93.53\% & \underline{86.61\%} & 71.65\% \\
  \hline
\end{tabular}
 }
\vspace{-2mm}
\label{sota}
\vspace{-3mm}
\end{table*}
\section{Experimental Setup}
\label{sec:experimental_setup}

\subsection{Implementation Details}

\noindent \textbf{a). Training Spatial ($\mathcal{A}^{s}$) and Temporal ($\mathcal{A}^{t}$) AE's:} We closely follow the training procedure described in \cite{astrid2021learning} to train $\mathcal{A}^{s}$ and $\mathcal{A}^{t}$. The architecture of $\mathcal{A}^{s}$ and $\mathcal{A}^{t}$ is adapted from \cite{Gong_2019_ICCV}, however instead of relying on single channel image as input we use all 3 channels.  $\mathcal{A}^{s}$ and $\mathcal{A}^{t}$ were trained on respective datasets
from scratch with the objective defined in eq.~\ref{spatial_loss_objective} and eq.~\ref{temporal_loss_objective} respectively on 2 NVIDIA GeForce 2080 Ti GPUs with effective batch size ($\mathcal{B}$) of 24 distributed across the GPUs (12 each). The input to  $\mathcal{A}^{s}$ and $\mathcal{A}^{t}$ is of  size $(\mathcal{B} \times \mathcal{T} \times 3 \times 256 \times 256)$, where $\mathcal{T}=16$.
The spatial and temporal   PAs were sampled by probability  $p_s=0.4$ and $p_t=0.5$ respectively. $\mathcal{A}^{s}$ is trained with Adam optimiser for 25 epochs with a learning rate of $10^{-4}$. During training, the reconstruction loss is calculated across all 16 frames of the sequence. The training of the $\mathcal{A}^{t}$ follows a similar procedure, however the input to the model is the optical flow representing normal events i.e  $\phi$.


\noindent \textbf{b). Training the Discriminator ($\mathcal{D}$):}
During the training phase, the input to $\mathcal{D}$ has a batch size of $16$ and  feature dimension of $512$. The model was trained using a SGD optimiser with a learning rate of $0.02$, momentum of $0.9$ and weight decay of $10^{-3}$ for $20$ epochs.  The groundtruth for normal and PAs samples are given labels $0$ and $1$ respectively. See section~\ref{add_details_supp} (supp.) for ViFi-CLIP \cite{rasheed2023fine} feature extraction and additional details. 

\noindent The complete  pipeline is depicted in Figure~\ref{fig:arch}.
\subsection{Inference}
During inference (Figure~\ref{fig:inference} in supp.), our goal is to measure all three types of anomaly indicators of all frames of the test video in the given dataset i.e reconstruction quality, temporal irregularity and semantic inconsistency.
Therefore, our anomaly score should holistically combine these aspects to gain deeper insights into real-world anomalies in videos.

In order to measure the reconstruction quality, we follow the recent works of \cite{9090171,liu2018future,park2020learning}, which utilise normalised Peak Signal to Noise Ratio $P_{t}$ (PSNR) between the test input frame at time $t$ and its reconstruction from $\mathcal{A}^{s}$ to calculate the anomaly score $\omega^{(t)}_1$. The input to $\mathcal{A}^{s}$ during inference is given in a sliding window fashion and has  dimensions $1 \times 16 \times 3 \times 256 \times 256$, where batch size is 1 and 16 represents number of frames. At test time, only the $9^{th}$ frame of a sequence is considered for anomaly score calculation as in \cite{astrid2021learning}. For measuring  temporal irregularity, a similar strategy is followed as for frames but instead of measuring the PSNR, the normalised $\mathcal{L}_2$ loss (denoted by $\omega^{(t)}_2$) is computed between the input test $\phi$ at time $t$ and its reconstruction from $\mathcal{A}^{t}$. For measuring  semantic inconsistency, the sequence of input frames is fed into $\mathcal{D}$ in a sliding window fashion with a window size of 16. We compute the output probability of a frame at time $t$ to be anomalous from its ViFi-CLIP feature representation and  denote it by $\omega^{(t)}_3$. The aggregate of anomaly score for all  three components is given by the following weighted average:


\begin{equation}
\label{anomaly_score_aggregation}
\mathbf{\omega}^{(t)}_{agg} =
\begin{cases}
    \mathbf{\eta}_1 \mathbf{\omega}^{(t)}_1 + \mathbf{\eta}_2 \mathbf{\omega}^{(t)}_2 + \mathbf{\eta}_3 \mathbf{\omega}^{(t)}_3, & \text{w/ } \mathcal{D} \\
    \mathbf{\eta}_1 \mathbf{\omega}^{(t)}_1 + \mathbf{\eta}_2 \mathbf{\omega}^{(t)}_2, & \text{w/o } \mathcal{D}; (\eta_3=0) \\
\end{cases}
\end{equation}

\noindent 
where $\eta_1$, $\eta_2$, $\eta_3$ are tuned for every dataset. (Refer to section~\ref{eval_criteria} in supp. material for further details)

\begin{figure*}
     \centering
     \includegraphics[width=0.88\textwidth]{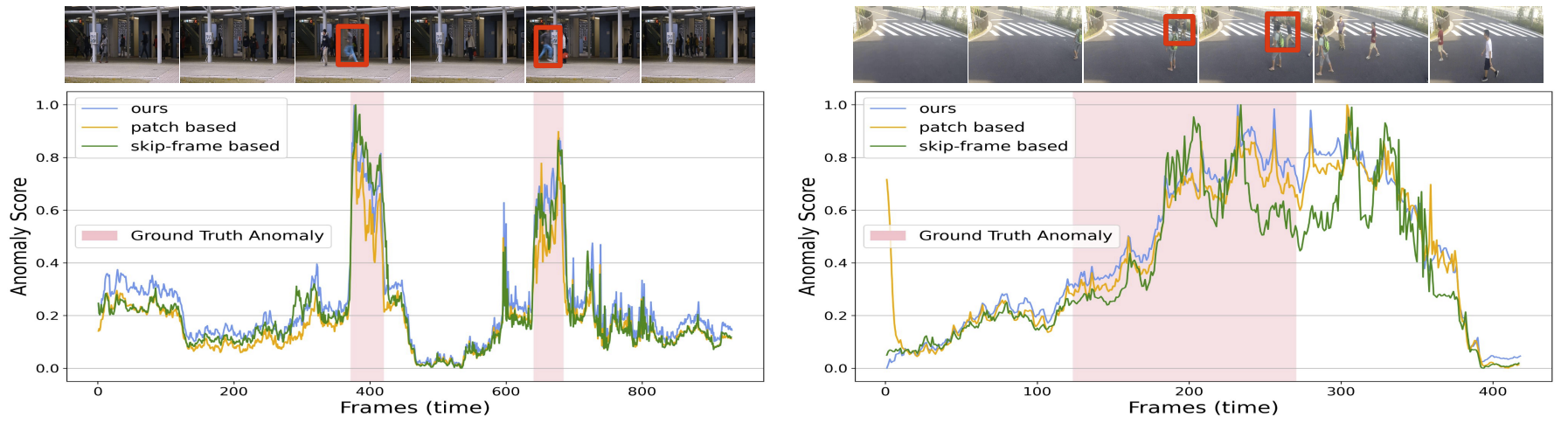}
     \caption{Qualitative Assessment : Visualisation of anomaly score over time for sample videos in Avenue (left) and ShanghaiTech (right).} 
     \label{fig:ano_score_avenue_shanghai}
\end{figure*}

\subsection{Results}
We performed extensive and exhaustive quantitative and qualitative assessments on four datasets namely  Ped2 \cite{ped2}, Avenue \cite{avenue_6751449}, ShanghaiTech \cite{luo2017revisit} and UBnormal \cite{Acsintoae_2022_CVPR}. 

\noindent \textbf{Baselines:} We compare our results with memory based AE \cite{gong2019memorizing,park2020learning} and other reconstruction based method trained with pseudo-anomalous samples created using other simulation techniques \cite{astrid2021learning,astrid2021synthetic}. The network trained without any PAs is represented as the standard \underline{\textit{baseline}}. The model design of the AE is fixed across all the experimental settings. Object-level information is only considered for perturbing the normal data during training while at inference we evaluate results strictly based on reconstruction and classification outputs. Hence our method is not directly comparable to object-centric methods.

\begin{table}[h]
\caption{Micro AUC score comparison between our approach and  existing state-of-the-art methods on val split of UBnormal \cite{Acsintoae_2022_CVPR}.}
\label{results_ubnormal}
    \centering
    \resizebox{\columnwidth}{!}{%
    \begin{tabular}{lc}
    \hline
        Reconstruction Methods & UBnormal \cite{Acsintoae_2022_CVPR} \\
        \hline 
        \textit{Baseline} (without PAs) & 54.06 \% \\
        LNTRA Astrid \etal \cite{astrid2021learning}  - Patch based      &  57.09 \% \\
        LNTRA Astrid \etal \cite{astrid2021learning}  - Skip-frame based      &  55.48 \% \\
        \rowcolor{Gray}
        Ours w/o $\mathcal{D}$ & \underline{57.53} \% \\
        \rowcolor{Gray}
        Ours w/ $\mathcal{D}$ & \textbf{57.98 \%} \\
        \hline 
    \end{tabular}
    }
    
\end{table}

\begin{figure*}[t]
      \centering
       \includegraphics[width=0.87\textwidth]
       {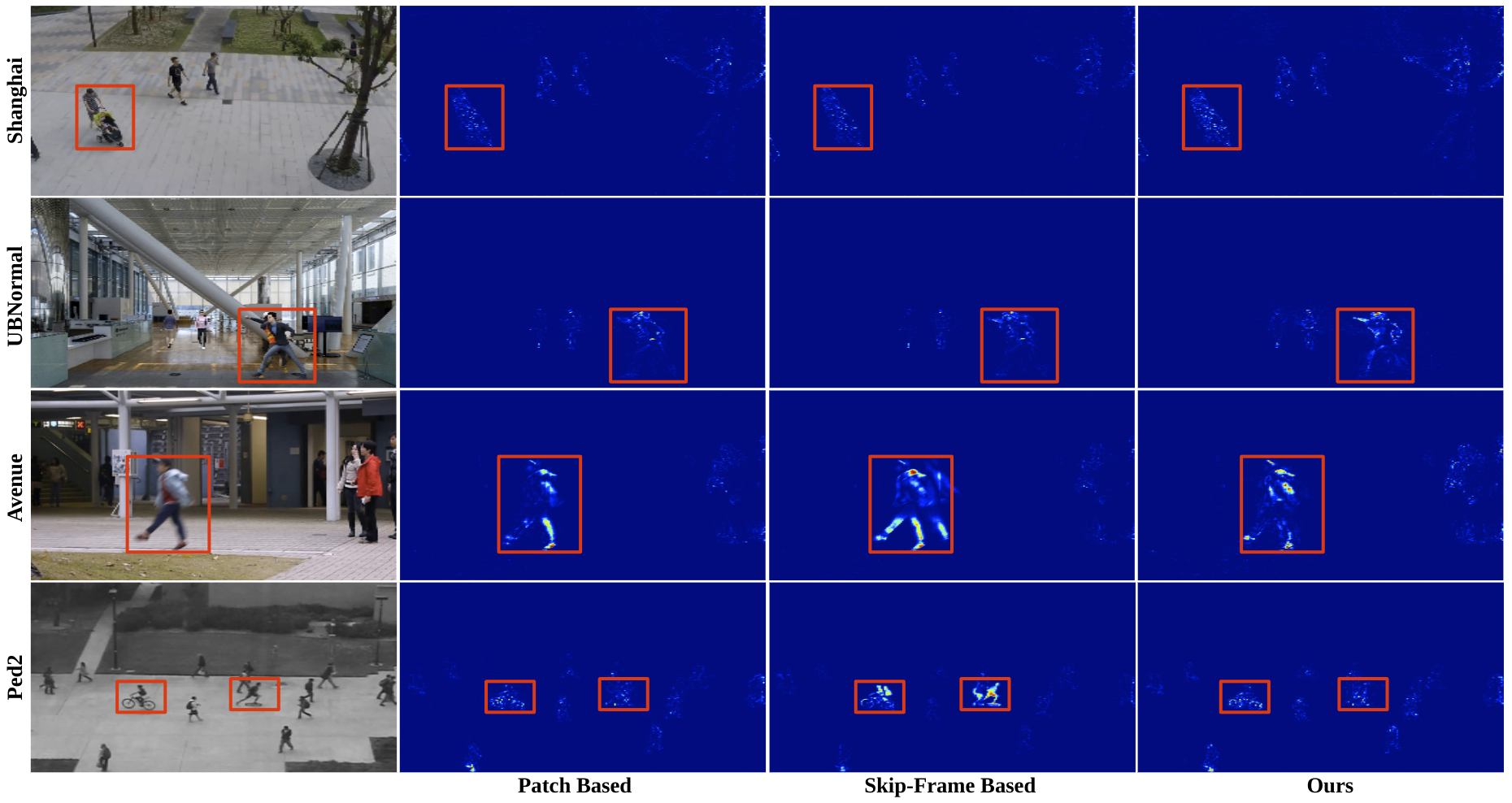}
      \caption{Visualisation of error heatmap for sample videos. Compared with other PAs generator methods in LNTRA~\cite{astrid2021learning}.
      } 
      \label{fig:error_heatmap_UBnormal_shanghai}
 \end{figure*}

\noindent \textbf{1. Quantitative Assessment:} In Table~\ref{results_sota}, we report  micro AUC comparisons of overall scores of our model and  existing state-of-the-art (SOTA) methods on test sets of Ped2, Avenue and Shanghai datasets. We follow the same practice as in \cite{astrid2021learning} of dividing the SOTA methods into 5 categories. Our method is closest to reconstruction based methods though we also avail the discriminator $\mathcal{D}$ as the auxiliary component to learn the distance between normal data distribution and PAs distribution. For clarity, we provide results \underline{\textit{with and without $\mathcal{D}$}} for all the datasets. 
Compared to  memory-based networks, our unified framework trained on synthetically generated spatio-temporal PAs outperforms MemAE \cite{gong2019memorizing} and MNAD-Reconstruction \cite{park2020learning} on Avenue and Shanghai while on Ped2 surpasses MNAD-Reconstruction and achieves comparable performance as MemAE. We also compare our results with other PAs generator methods such as STEAL Net \cite{astrid2021synthetic} and LNTRA \cite{astrid2021learning}. We observe that on the Avenue dataset our model outperforms LNTRA (patch, skip-frame based) though marginally lags behind STEAL-Net whereas STEAL-Net and LNTRA achieve better performance than our model on Ped2 and Shanghai dataset. However such methods generate PAs under bold assumptions and inductive biases which may cause them to fail in particular cases. We report such cases in the Ablation study (Figure~\ref{fig:error_heatmap_UBnormal_shanghai}). We also show in Table \ref{ablation_transfer} that the transfer performance of our model is on par with other PAs generation methods (see section \ref{ablation_section}). We do employ optical flow like other methods (e.g~ Frame-Pred \cite{liu2018future}) and observe that our results outperform Frame-Pred on the Avenue, achieve comparable performance on ShanghaiTech and are marginally less on Ped2.


In Table~\ref{results_ubnormal}, we show a comparison between baseline, LNTRA and our approach on the validation set of the UBnormal dataset using only the normal videos in the training split. This is done to ensure consistency in evaluation under the OCC setting (refer to section~\ref{dataset_desc} in supp for data-split details). The training and evaluation for baseline and LNTRA (patch, skip-frame) based methods on UBnormal was performed using scripts provided by the authors of LNTRA\footnote{\href{https://github.com/aseuteurideu/LearningNotToReconstructAnomalies}{https://github.com/aseuteurideu/LearningNotToReconstructAnomalies}}. We observe that our method outperforms baseline and LNTRA achieving micro AUC score of 57.98\% and  implying that our PAs are generic and applicable for more diverse anomalous scenarios. Both in Table \ref{results_sota}, \ref{results_ubnormal} we notice that the effect of adding $\mathcal{D}$ is minimal, which validates the intuition that VAD cannot be directly addressed as a classification problem.


Table \ref{results_sota}, \ref{results_ubnormal} shows that no single reconstruction-based method excels on all datasets. This is because anomalies are context-dependent. Different methods have inductive biases that work for specific datasets but not others. Our work provides a generic solution towards generating PAs without making bold assumptions about dataset's anomalies.

\noindent \textbf{2. Qualitative Assessment:}  
We conduct qualitative analysis of the anomaly score over time for sample videos in Avenue, Shanghai (Figure~\ref{fig:ano_score_avenue_shanghai}) and Ped2, UBnormal (Figure~\ref{fig:ano_score_ped2_ubnormal} in supp). We also compare our model's anomaly score over time with those obtained from LNTRA skip-frame and patch-based methods. It can be concluded that  on the Avenue and Ped2 datasets, our method detects anomalies fairly well and performance is equivalent with LNTRA models. Though there exist certain failure cases in the Shanghai and UBnormal datasets which occur due to anomalies occurring due to abnormal interaction between two objects i.e. fighting between two individuals in Shanghai and accident with a bike in UBnormal. Even though our PAs generator is generic it fails to emulate such complex real-world anomalies.

\subsection{Ablation Studies}
\label{ablation_section}
\noindent \textbf{1: How transferable are PAs?}
We also examine how well PAs transfer across various VAD datasets. We use our pre-trained model on UBnormal dataset, which contains a wide range of anomalies and backgrounds, making it suitable for transferability. We tested the model on rest of the datasets without fine-tuning. Our results in Table~\ref{ablation_transfer} show that our model outperforms the patch-based method on all other datasets while achieves competitive performance compared to the skip-frame based method. This provides an interesting insight that our PAs are generic and transferable.


\begin{table}[h]
\caption{Transfer Performance : micro-AUC scores.}
\label{ablation_transfer}
    \centering
    \begin{tabular}{lccc}
    \hline
        Method & Ped2 & Avenue & Shanghai \\
    \hline
        Patch \cite{astrid2021learning}      &  78.80 \%  & 43.94 \% & 61.57 \% \\
        Skip-Frame \cite{astrid2021learning}      &  85.21 \% & 83.82 \% & 70.52 \% \\
        \rowcolor{Gray}
        Ours w/ $\mathcal{D}$     &  85.37 \% & 83.50 \% & 70.07 \% \\
        \hline 
    \end{tabular}
    
\end{table}


\noindent \textbf{2: How to interpret PAs?}
In Figure~\ref{fig:error_heatmap_UBnormal_shanghai}, we compare error heatmaps generated using a model trained with patch and skip-frame based PAs and with our spatial-PAs on all the respective datasets. Since skip-frame and patch based PAs carry strong assumptions, they tend to have problems detecting complicated real-world anomalies in ShanghaiTech such as a baby carriage (anomalous object) whereas our model trained with spatial-PAs yields high error for such cases. Furthermore, our PAs also give strong results on the synthetic dataset UBnormal, where patch and skip-frame based PAs fail to detect complex violent scenes as temporal irregularity induced through skip-frames is not generic. However, even our spatial-PAs, which are not explicitly trained to detect temporal anomalies are able to determine such real-world anomalies. On Avenue and Ped2 datasets, our model gives comparable error to patch based PAs for an anomalous activity however we observe that skip-frame based PAs overly estimates the reconstruction error for the same. Intuitively this indicates that even though skip-frame performs reasonably well on benchmark datasets but it is susceptible to amplification of the error. An explanation for this phenomena could be due to underlying strong assumption of skipping frames based on a specific stride value to model temporal irregularity. These observations validate that our PAs are generalised and enable understanding of which real-world anomalies can be detected using which type of PAs.



\noindent \textbf{3: Random vs Segmentation masks:}
Table \ref{ablation_mask} shows the effect of using random and segmentation masks for generating spatial PAs. We observe that using a segmentation mask gives better AUC score on Ped2 and Avenue dataset, which is intuitively justified as segmentation masks contain more semantic information. Despite this, our method is flexible in terms of type of mask chosen.

\begin{table}[h!]
\caption{Effect of Random and Segmentation masks on micro-AUC scores, using the output of $\mathcal{A}^{s}$ when trained with $p_s = 0.4$.}
\label{ablation_mask}
    \centering
    \begin{tabular}{ccc}
    \hline
        Mask Type &  Ped2 & Avenue \\
    \hline
        Random Mask      &  91.18 \%  & 83.13 \% \\
        Segmentation Mask      &  92.71 \% & 84.51 \% \\
        \hline 
    \end{tabular}  
\end{table}
\section{Conclusions and Discussion}
\label{sec:conclusion}
In this paper we presented a novel and generic spatio-temporal PAs generator vital for VAD tasks without incorporating strong inductive biases. We achieve this by adding perturbation in the frames of normal videos by inpainting a masked out region in the frames using a pre-trained LDM and by distorting optical flow by applying mixup-like augmentation (Figure~\ref{fig:pseudo_ano}). We also introduced a unified VAD framework that learns three types of anomaly indicators i.e. reconstruction quality, temporal irregularity and semantic inconsistency in an OCC setting (Figure~\ref{fig:arch}). Through extensive evaluation, we show that our framework achieves on par performance with other SOTA reconstruction methods and PA generators with predefined assumptions across multiple datasets (Table~\ref{results_sota}, \ref{results_ubnormal}) indicating the effectiveness, generalisation and transferability of our PAs.

There are limitations with this work. First, our model was not trained in an end-to-end fashion due to limited computational resources available. It will be interesting to make this setting adaptive in nature by learning a policy network to select which anomaly indicator among poor reconstruction quality, temporal irregularity and semantic inconsistency contributes more towards detection of  real-world anomalies. Second, the notion of generating latent space PAs for VAD through LDMs or manifold mixup remains to be investigated. Third, in this work micro AUC scores are used for evaluation though the method needs to be further validated on other metrics such as region, tracking based detection criteria. These limitations will be addressed in our future work.

\section{Acknowledgement}
This research is supported by Science Foundation Ireland under the Grant Number SFI/12/RC/2289\_P2, funded by the European Regional Development Fund and Tobii FotoNation. Dedicated to Dr. Kevin McGuinness, with lasting gratitude for his invaluable contributions and enduring influence.
\clearpage
\setcounter{page}{1}
\maketitlesupplementary




\section{Datasets}
\label{dataset_desc}
\noindent \textbf{Ped2} \cite{ped2} dataset comprises of 16 training and 12 test videos and all videos have the same scene in the background. The videos with normal events consist of pedestrians only, whereas the videos with anomalous events include bikes, skateboards and carts apart from pedestrians.

\noindent \textbf{Avenue} \cite{avenue_6751449} dataset comprises of 16 training and 21 test videos with every video having the same background scene. Normal events involve people routinely walking around while the abnormal instances include abnormal objects such as bikes and abnormal human actions such as unusual walking directions, running around or throwing things.

\noindent \textbf{ShanghaiTech} \cite{luo2017revisit} dataset includes 330 training and 107 test videos recorded at 13 different background locations with complex lightning conditions and camera angles, making it the one of the largest one-class anomaly detection datasets. The test split captures a total of 130 anomalous events including running, riding a bicycle and fighting.

\noindent \textbf{UBnormal} \cite{Acsintoae_2022_CVPR} is a synthetic dataset with multi-scene backgrounds and a diverse set of anomalies. The dataset consists of training, validation and test split with both normal and abnormal events. The normal events include walking, talking on the phone, walking while texting, standing, sitting, yelling and talking with others. It is to be noted that abnormal events in each of the train, validation and test split are different to each other. The train split includes abnormal events like \textit{falling, dancing, walking injured, running injured, crawling, and stumbling walk}. The validation split comprises \textit{fighting, sleeping, dancing, stealing, and rotating 360 degrees}. All the evaluations are conducted on the validation set.


\noindent \textbf{UBnormal data-split under OCC Setting. } In order to use this dataset in the one class classification (OCC) setting, we train our model using only the normal 186 videos in the training split and the pseudo-anomalies (PAs) generated using them (i.e. totally ignoring the abnormal samples provided in the train set). We tested our model on all the videos in the validation split, comprising of 64 videos with both normal and abnormal events. 
Such a setting was chosen to keep consistency in evaluation as with other datasets under the OCC setting. The frame-level groundtruth annotation for validation set of UBnormal \cite{Acsintoae_2022_CVPR} was created using the script\footnote{\href{https://github.com/lilygeorgescu/UBnormal/tree/main/scripts}{https://github.com/lilygeorgescu/UBnormal/tree/main/scripts}} provided by the authors.


\begin{figure}
    \centering
    \includegraphics[width=0.5\textwidth]{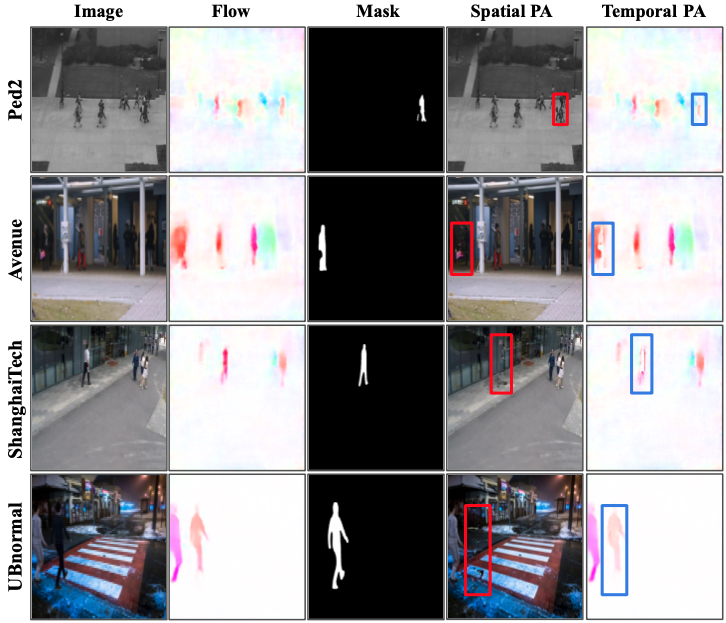}
    \caption{Qualitative Assessment : Visualisation of spatial and temporal PAs for all  4 datasets. Here we only show segmentation masks however the approach also works with random masks.} 
    \label{fig:pseudo_ano2}
\end{figure}

\section{Additional Details and Insights}
\label{add_details_supp}

\noindent \textbf{1: Pseudo-Anomaly Construction. } 
We take an off-the-shelf Latent Diffusion Model \cite{rombach2022high} (LDM\footnote{\href{https://github.com/CompVis/latent-diffusion/tree/main}{https://github.com/CompVis/latent-diffusion/tree/main}})  pre-trained on the Places dataset \cite{zhou2017places}. We do not perform any finetuning of the LDM on any video anomaly dataset and therefore it is ``under-trained" on video data and hence capable of spatially distorting them. For inpainting the masked out regions of the images, $50$ steps of inference were carried out. It is to be noted that due to lack of computational resources we did not experiment with other values of timesteps. A very low number of timesteps may produce mostly noisy inpainting output while a very high value  might result in inpainted images very close to the input image. The strategy for generation of random and segmentation masks was adopted from the code\footnote{\href{https://github.com/advimman/lama/tree/main/saicinpainting/evaluation/masks}{https://github.com/advimman/lama/tree/main/saicinpainting}} provided by the authors of LAMA \cite{suvorov2022resolution}. If segmentation mask was detected for a frame, a random mask was selected instead. Figure~\ref{fig:pseudo_ano2} depicts more examples of generated PAs.




\noindent \textbf{2: Extracting ViFi-CLIP Features.} 
For the training split of the benchmark datasets and their corresponding spatial pseudo-anomalies, we extract frame level features using the ViFi-CLIP \cite{rasheed2023fine} model. The input to the ViFi-CLIP model has  size : $\mathcal{B}^\prime \times \mathcal{T}^\prime \times 3 \times 224 \times 224$, where $\mathcal{B}^\prime$ (batch size) was set to $1$ and $\mathcal{T}^\prime$ (\# of frames) was set to $16$. All  frames were passed into ViFi-CLIP in a sliding window fashion with a stride of $16$ therefore we obtain a $512$-dimensional feature for every frame. ViFi-CLIP uses the backbone of ViT-B/16 \cite{dosovitskiy2020image} and is pre-trained on Kinetics-400 \cite{kay2017kinetics}.  It is to be noted that the ViFi-CLIP model performs temporal pooling of the CLIP \cite{radford2021learning} features, however we do not perform temporal pooling and use the frame level representations as during inference we evaluate our pipeline using frame level micro AUC scores. For the frames of the videos in test split (Ped2, Avenue, ShanghaiTech) and validation split (UBnormal), we follow the same procedure for feature extraction. 

\noindent \textbf{3: Effect of changing the probability of sampling PAs.} 
We conduct an experimental study by varying the probability of sampling spatial and temporal PAs ($p_s$, $p_t$) on Ped2 during training between 0.1 to 0.5 and measuring micro AUC scores during inference. Figure~\ref{fig:pa_prob_supp} shows that the model achieves best performance when $p_s = 0.4$ and $p_t = 0.5$.

\noindent \textbf{4: Inference Time. } The average inference time calculated over three runs for a single frame on a single Nvidia RTX-2080-Ti GPU is 123.35ms.

\begin{figure}
    \centering
    \includegraphics[width=0.5\textwidth]{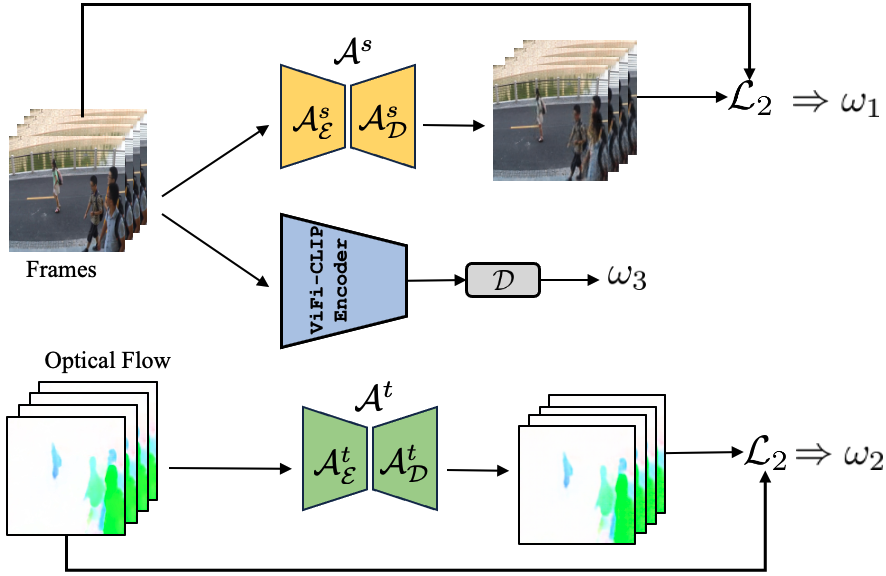}
    \caption{During inference, aggregate anomaly score is computed by calculating the weighted sum (eq \ref{anomaly_score_aggregation}) of all the three types of anomaly information; reconstruction quality $\omega_1$ (eq \ref{reconstruction_quality_score}), temporal irregularity $\omega_2$ (eq \ref{temporal_regularity_score}) and semantic inconsistency $\omega_3$.}
    \label{fig:inference}
\end{figure}

\section{Evaluation Criteria}
\label{eval_criteria}
To measure the reconstruction quality, we follow the recent works of \cite{9090171,liu2018future,park2020learning}, which utilised normalized Peak Signal to Noise Ratio (PSNR) $P_{t}$ between an input frame and its reconstruction to calculate the anomaly score. This is illustrated in the following equation.

\begin{equation}
    P_t = \text{10 } \text{log}_{10} \frac{M^{2}_{\mathbf{\hat{x}_{t}}}}{\frac{1}{R} || \mathbf{\hat{x}_{t}} - \mathbf{x_{t}} ||^2_{2} },
\end{equation}

\begin{equation}
    \label{reconstruction_quality_score}
    \mathbf{\omega}^{(t)}_1 = 1 - \frac{P_{t} - \text{min}_{t} (P_{t}) }{\text{max}_{t} (P_{t}) - \text{min}_{t} (P_{t})},
\end{equation}

\noindent 
where $\mathbf{x_{t}}$ is the input frame at time $t$,   $\mathbf{\hat{x}_{t}}$ represents reconstruction of $\mathbf{x_{t}}$, $R$ denotes the total number of pixels in $\mathbf{\hat{x}_{t}}$ and $M_{\mathbf{\hat{x}_{t}}}$ is the maximum possible pixel value of $\mathbf{\hat{x}_{t}}$. The anomaly score $\mathbf{\omega}^{(t)}_1$ is an indicator of reconstruction quality of the input frame. For measuring the temporal irregularity, we compute the normalised $\mathcal{L}_2$ loss between input optical flow at time $t$ and its reconstruction given by the equation:

\begin{equation}
    \label{temporal_regularity_score}
    \mathbf{\omega}^{(t)}_2 = \frac{1}{R^\prime} || \hat{\phi}(\mathbf{x_{t}},\mathbf{x_{(t+1)}}) - \phi(\mathbf{x_{t}},\mathbf{x_{(t+1)}}) ||^{2}_{2}, 
\end{equation}

\noindent 
where $\phi(\mathbf{x_{t}},\mathbf{x_{(t+1)}})$ is the input optical flow frame calculated using consecutive frames $\mathbf{x_{t}}$ and $\mathbf{x_{(t+1)}}$, $\hat{\phi}(\mathbf{x_{t}},\mathbf{x_{(t+1)}})$ represents the reconstruction of $\phi(\mathbf{x_{t}},\mathbf{x_{(t+1)}})$, $R^\prime$ denotes the total number of pixels in $\hat{\phi}(\mathbf{x_{t}},\mathbf{x_{(t+1)}})$. To measure the semantic inconsistency, the input frames sequence is fed into $\mathcal{D}$ in a sliding window fashion with a window size of 16. The output probability ($\omega^{(t)}_3$) of a frame at time $t$ to be anomalous is computed using its ViFi-CLIP feature representation. 

A higher value of $\mathbf{\omega}^{(t)}_1$, $\mathbf{\omega}^{(t)}_2$ and $\mathbf{\omega}^{(t)}_3$   represents higher reconstruction error for frame  and optical flow and high anomaly probability at time $t$ in the test videos during inference. Alternatively, they are indicators of poor reconstruction quality, temporal irregularity and semantic inconsistency and their aggregation can aid in determining real-world anomalies. The aggregate anomaly score is given by:

\begin{align*}
\label{anomaly_score_aggregation_supp}
\mathbf{\omega}^{(t)}_{agg} =
\begin{cases}
    \mathbf{\eta}_1 \mathbf{\omega}^{(t)}_1 + \mathbf{\eta}_2 \mathbf{\omega}^{(t)}_2 + \mathbf{\eta}_3 \mathbf{\omega}^{(t)}_3, & \text{w/ } \mathcal{D} \\
    \mathbf{\eta}_1 \mathbf{\omega}^{(t)}_1 + \mathbf{\eta}_2 \mathbf{\omega}^{(t)}_2, & \text{w/o } \mathcal{D}; (\eta_3=0) \\
\end{cases} 
\end{align*}

\noindent 
where $\mathbf{\eta}_1$, $\mathbf{\eta}_2$, $\mathbf{\eta}_3$ are weights assigned to $\mathbf{\omega}^{(t)}_1$, $\mathbf{\omega}^{(t)}_2$ and $\mathbf{\omega}^{(t)}_3$ respectively. The values of $\eta_1$, $\eta_2$ and $\eta_3$ lies in the interval $[0,1]$ and their sum is equal to 1.
We manually tune the values of $\eta_1$, $\eta_2$, $\eta_3$ for all the datasets. The values of ($\eta_1,\eta_2,\eta_3$) for all the datasets are given by - Ped2 (0.65,0.25,0.1), Avenue (0.45,0.5,0.05), Shanghai (0.85, 0.13, 0.02) and UBnormal (0.4, 0.5, 0.1). In all of the cases, any of the three component can be excluded during evaluation by setting the corresponding weight ($\eta_1,\eta_2,\eta_3$) to zero. \textbf{Note :} We also experimented with the learnt weights for the three anomaly indicators but there was a marginal decrease in the performance compared to manually tuning their weights.

\noindent \textbf{Evaluation Metric. }
For evaluation, we follow the standard metric of frame-level area under the ROC curve (micro-AUC) as in \cite{zaheer2020old}. We obtain the ROC curve by varying the anomaly score thresholds to plot False Positive Rate and True Positive Rate for the whole test set for a given dataset. Higher AUC values indicate better performance and more accurate detection of anomalies.



\begin{table*}[h!]
 \caption{Discriminator $\mathcal{D}$ architecture details}
 \vspace{0.2em}
 \label{disc_arch}
    \centering
     \resizebox{0.35\linewidth}{!}{
    \begin{tabular}{c|c}
        \hline 
        \hline 
             Layers & (Input size, Output size) \\
        \hline
        Linear Layer 1 & (512,128) \\
        \hline
        ReLU & - \\
        \hline
        Linear Layer 2 & (128,1) \\
        \hline
        \hline 
    \end{tabular}
   }
\end{table*}

\begin{table*}[h!]
\caption{Autoencoder ($\mathcal{A}^{s}$ and $\mathcal{A}^{t}$) architecture details}
\vspace{0.5em}
\label{model_arch}
    \centering
    \resizebox{16cm}{!}{%
    \begin{tabular}{c|c|c|c|c|c|c|c}
    \hline
    \hline
    & Layer & Input Channels & Output Channels & Filter Size & Stride & Padding & Negative Slope \\
    \hline
    \multirow{12}{*}{Encoder} & Conv3D &  3 & 96 & (3,3,3) & (1,2,2) & (1,1,1) & - \\
                             & BatchNorm3D & - & - & - & - & - & - \\
                             & LeakyReLU & - & - & - & - & - & 0.2 \\

                             & Conv3D & 96 & 128 & (3,3,3) & (2,2,2) & (1,1,1) & - \\
                             & BatchNorm3D & - & - & - & - & - & - \\
                             & LeakyReLU & - & - & - & - & - & 0.2 \\

                             & Conv3D & 128 & 256 & (3,3,3) & (2,2,2) & (1,1,1) & - \\
                             & BatchNorm3D & - & - & - & - & - & - \\
                             & LeakyReLU & - & - & - & - & - & 0.2 \\

                             & Conv3D & 256 & 256 & (3,3,3) & (2,2,2) & (1,1,1) & - \\
                             & BatchNorm3D & - & - & - & - & - & - \\
                             & LeakyReLU & - & - & - & - & - & 0.2 \\
    \hline
    \multirow{12}{*}{Decoder} & ConvTranspose3D & 256 & 256 & (3,3,3) & (2,2,2) & (1,1,1) & - \\
                             & BatchNorm3D & - & - & - & - & - & - \\
                             & LeakyReLU & - & - & - & - & - & 0.2 \\

                             & ConvTranspose3D & 256 & 128 & (3,3,3) & (2,2,2) & (1,1,1) & - \\
                             & BatchNorm3D & - & - & - & - & - & - \\
                             & LeakyReLU & - & - & - & - & - & 0.2 \\

                             & ConvTranspose3D & 128 & 96 & (3,3,3) & (2,2,2) & (1,1,1) & - \\
                             & BatchNorm3D & - & - & - & - & - & - \\
                             & LeakyReLU & - & - & - & - & - & 0.2 \\

                             & ConvTranspose3D & 96 & 3 & (3,3,3) & (1,2,2) & (1,1,1) & - \\
                             & Tanh & - & - & - & - & - & - \\
    \hline
    \hline
    \end{tabular}
    }
\end{table*}

\begin{figure*}[h!]
     \centering
     \includegraphics[width=1.0\textwidth]{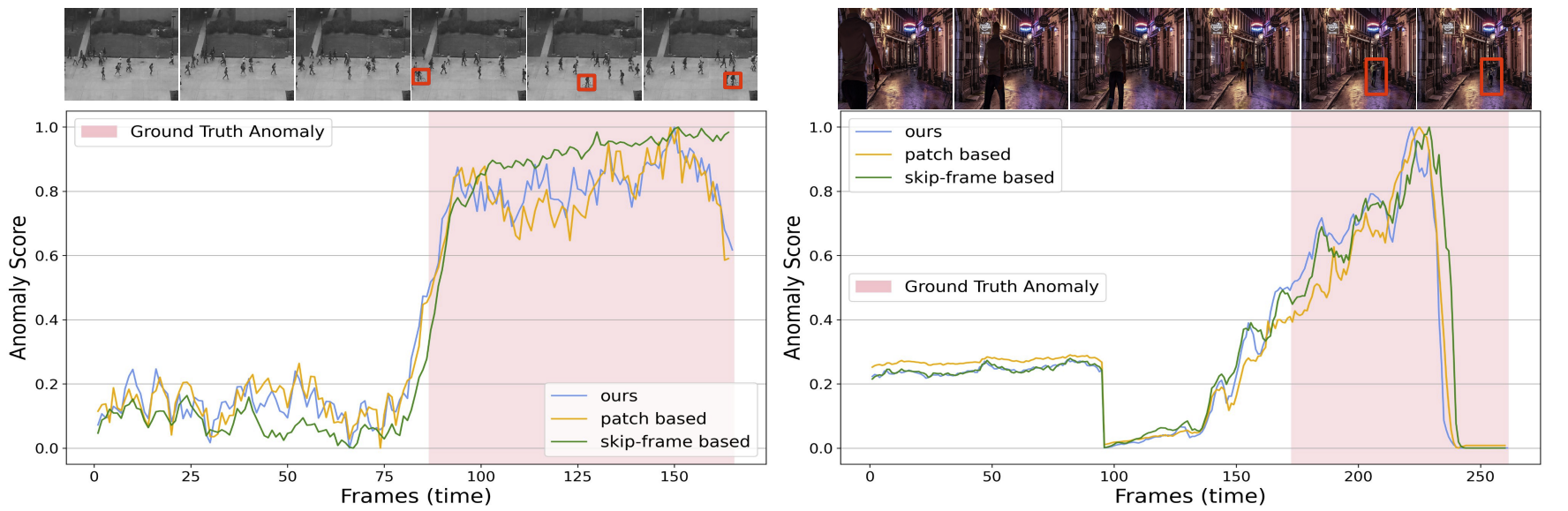}
     \vspace{0.2em}
     \caption{Qualitative Assessment : Visualization of anomaly score over time for sample videos in Ped2 (left) and UBnormal (right). Compared with other PAs generator and reconstruction based methods in LNTRA \cite{astrid2021learning} - patch and skip-frame based.} 
     \label{fig:ano_score_ped2_ubnormal}
\end{figure*}

\begin{figure*}[h!]
    \centering
    \begin{subfigure}{0.5\textwidth}
        \centering
        \includegraphics[width=1.0\textwidth]{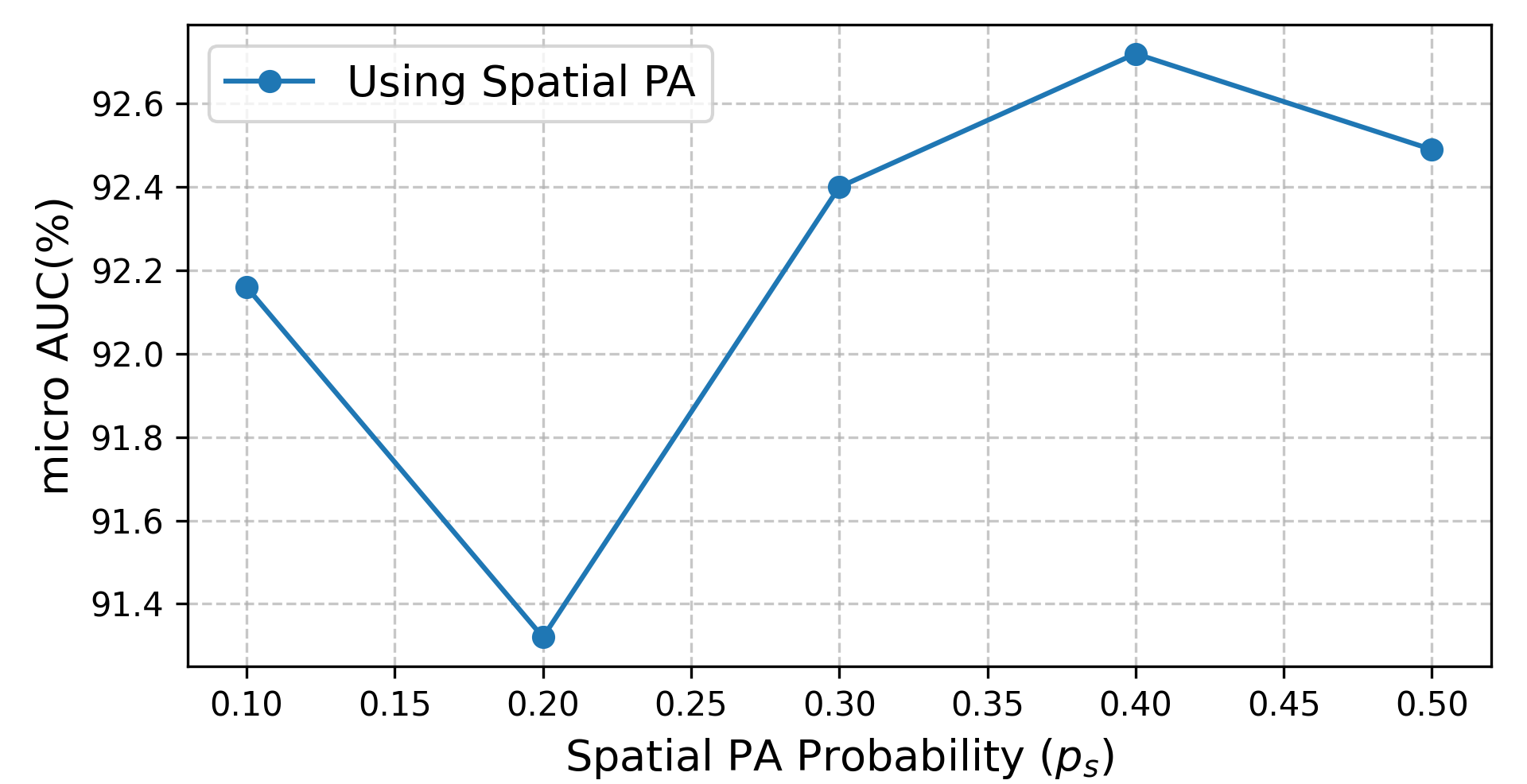}
        \caption{} 
    \end{subfigure}%
    \begin{subfigure}{0.5\textwidth}
        \centering
        \includegraphics[width=1.0\textwidth]{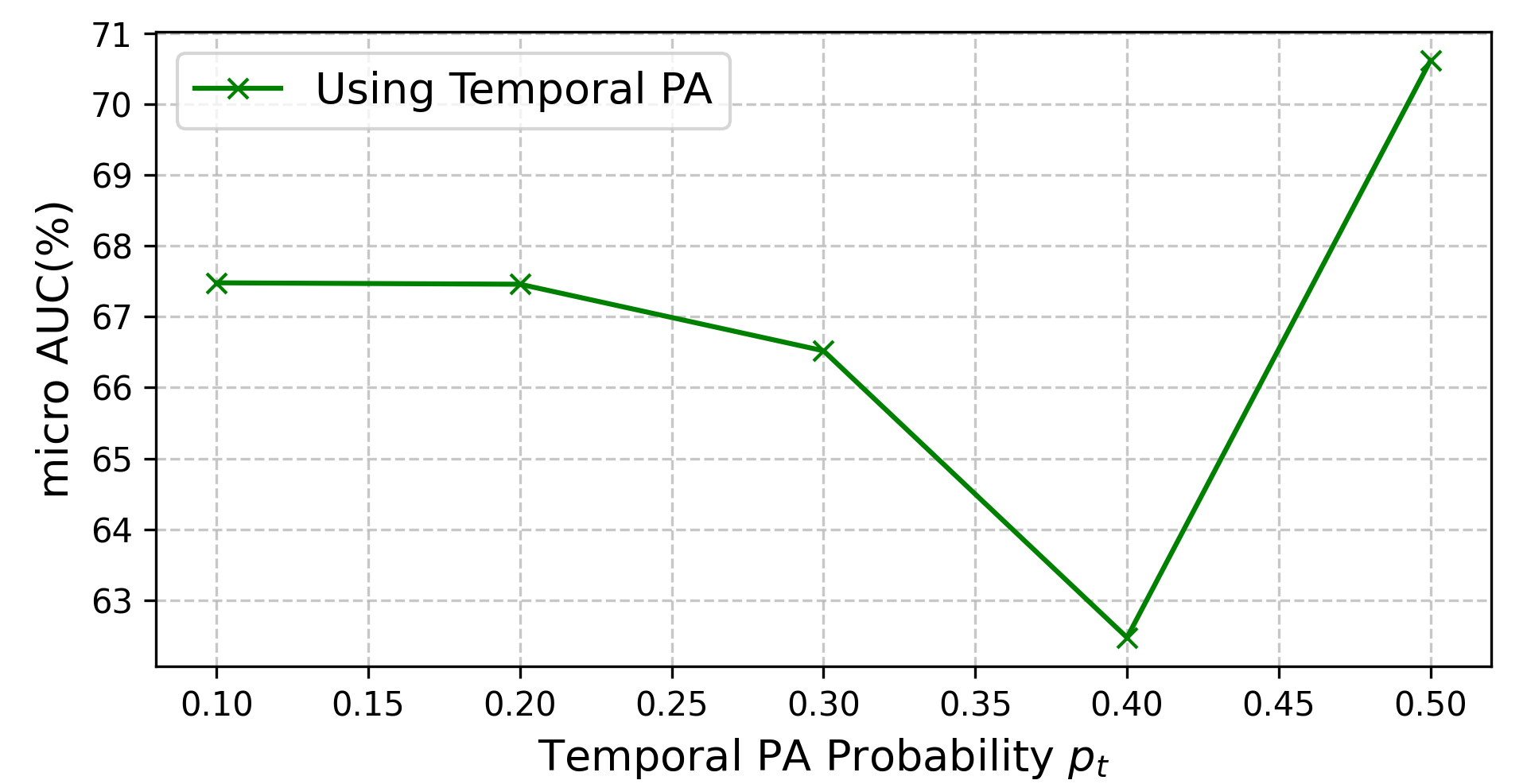}
        \caption{} 
    \end{subfigure}
    \caption{Comparison of micro-AUC scores on Ped2 dataset calculated from output of $\mathcal{A}^{s}$ ($\mathcal{A}^{t}$) trained on a range of values of $p_s$ ($p_t$) between \{0.1,0.5\}. We observe that setting $p_s = 0.4$ and $p_t = 0.5$ yields the best performance as shown in (a) and (b) respectively. These probability values are fixed for all other experiments.}
    \label{fig:pa_prob_supp}
\end{figure*}

{
    \small
    \bibliographystyle{ieeenat_fullname}
    \bibliography{main}
}


\end{document}